MaterialBENCH: Evaluating College-Level Materials Science Problem-Solving Abilities of Large Language Models

Michiko Yoshitake (1), Yuta Suzuki (2), Ryo Igarashi (1), Yoshitaka Ushiku (1), Keisuke Nagato (3)
((1) OMRON SINIC X, (2) Osaka Univ., (3) Univ. Tokyo)

Abstract

MaterialBENCH, a college-level benchmark dataset for large language models (LLMs) in the field of materials science, has been developed. This dataset consists of problem-answer pairs, derived from university textbooks and includes two types of problems: free-response and multiple-choice. In the multiple-choice format, three incorrect answers are added as options alongside the correct answer, requiring LLMs to select one of the four choices. Most problems are shared between the free-response and multiple-choice formats, differing only in the answer structure. We conducted experiments using MaterialBENCH with various LLMs, including ChatGPT-3.5, ChatGPT-4, Bard (at the time of the experiments), and GPT-3.5 and GPT-4 assessed via the OpenAI API. We analyzed and discussed the differences and similarities in performance across these LLMs as measured by MaterialBENCH. Additionally, we examined performance differences between the free-response and multiple-choice formats within the same models, as well as the influence of using system massages on multiple-choice problems. We anticipate that MaterialBENCH will inspire further advancements in LLMs' reasoning abilities, enabling them to solve more complicated problems and eventually contributing to materials research and discovery.

1 Introduction

Following the groundbreaking introduction of the transformer model with self-attention [1], many large language models (LLMs) based on this architecture have been rigorously developed [2,3,4].  Recent releases of generative artificial intelligence powered by LLMs, such as GPT-3.5, GPT-4, Bard (now Gemini), and Glaude 3, feature vast numbers of parameters, with their ability to respond to queries reaching or even surpassing human-level performance in some respects. To evaluate the performance of these LLMs, various benchmarks have been developed. At the time of the release of BERT [2], the GLUE benchmark [5, 6] was widely used. As many domain-specific BERT models emerged, different datasets tailored to specific fine-tuning tasks were employed, often drawn from scientific papers in related domains. In the materials science field, several BERT variants, including MatSciBERT [7], MatBERT [8], MaterialBERT [9], and BatteryBERT [10], have been developed and evaluated using distinct datasets specific to their respective areas. The performance of LLM models has been evaluated using the MMLU benchmark [11], which consists of multiple-choice questions from a broad spectrum of high-school and college subjects, as well as licensure exams. At the time of MMLU publication, the correct answer rate for college level mathematics, physics and chemistry was almost 0.25, which corresponds to random guessing on multiple-choice questions. This prompted the development of SciBench [12], a benchmark specifically targeting college level mathematics, physics, and chemistry. Recently, ChemBench [13] was developed, focusing on molecular chemistry (organic materials), with benchmark sentences drawn largely from safety-related topics such as toxicology and chemical safety. We have developed a benchmark dataset in the materials science field, where material properties such as hardness (mechanical property), dielectric constant (electric property), and magnetic susceptibility (magnetic property) are important. These properties originate from materials as aggregates of atoms or molecules, making the state of aggregation a key factor. Consequently, MaterialBENCH is very different from ChemBench, with many problems centered on inorganic materials (metals, semi-conductors, oxides, and so forth). In addition to evaluating the performance of existing LLMs in materials science, MaterialBENCH might be useful for evaluating user-modified models. Both open and closed LLMs provide options for tailoring models to specific domains. It should be noted that in SciBench [12], prompt techniques such as the chain of thought (CoT) and the use of plug-ins for outside tools such as Python were tested. However, many current LLMs use CoT techniques automatically, and some even incorporate plug-in tools as standard features.

## 2 The MaterialBench Dataset

To benchmark the capabilities and analyze the limitations of existing large language models (LLMs) in solving materials science problems, we have compiled a new dataset drawn from university-level textbooks that cover a broad spectrum of materials science. This section details the dataset construction process.

### 2.1 Textbook Selection

To ensure comprehensive coverage of materials science, we selected university-level textbooks for students majoring in the field. Since scientific principles underlying material properties are consistent across different types of materials, such as inorganic and organic compounds, we selected textbooks with such a wide viewpoint. Textbooks specializing in specific materials, such as "polymers", or specific phenomena, such as "magnetism", were excluded.

To avoid easy access and learning by LLMs from websites, we chose textbooks that are not readily accessible online and that are difficult to extract or transform into text. Specifically, we selected textbooks provided in PDF format rather than XML-type formats, as the latter are easier to parse. At the time of evaluation of LLMs with MaterialBENCH, PDF format posed challenges for learning, since the correct answers are typically located separately from the problems, at the end of books, making it difficult for models to learn the correct answers.

Accordingly, we selected two textbooks [14, 15] that have been extensively used in university courses as the open textbook from the materials science field.

## 2.2 Problem Selection

We meticulously collected each problem from the original textbooks in PDF format. Problems without answers provided by the authors of the textbooks were excluded from the benchmark selection. Additionally, redundant problems, such as those that only differed in materials names or compositions but shared the same solution method, were also excluded. The problems were manually collected by a human expert. To assess the impact of problem difficulty on LLM performance, we included a range of problems, from relatively simple knowledge or calculation-based questions to more complex problems requiring intricate combinations of calculations of equations or involving exceptionally small or large numbers. Some problems can be solved with a simple equation, allowing us to evaluate the ability of LLMs to find an appropriate equation to use.

The problems cover a wide array of topics, including atomic bonding, defects, diffusion, dislocation, strength, fracture, corrosion, electrochemistry, and electronic-, optical-, thermal-, magnetic properties. They also span materials such as metals, semiconductors, ceramics, polymers, and composites. However, phase diagrams and phase transitions were excluded from this benchmark due to the near-necessity of inputting figures, which introduces the separate challenge of figure recognition.

In total, 164 problems were selected, each with answers provided by the textbook authors. These answers were further verified by an expert to ensure accuracy, even though most textbooks did not include detailed solution processes. Among the 164 problems, 20 problems were excluded from the free-response section because their answers could not be unambiguously determined correct without choices for these 20 problems. Therefore, the remaining 144 problems were used for free-response problems and 164 for multiple-choice problems.

For the multiple-choice problems, an expert crafted three incorrect answers for each problem, designed by altering parts of the equations used in the solutions, reflecting common student mistakes. For problems with multiple correct answers, we either selected one or provided a correct choice as a pair of multiple answers. The selected problems demand a solid understanding of domain-specific knowledge, strong reasoning skills, adept calculation abilities, and the ability to comprehend complex concepts. Our dataset aims to present a comprehensive range of university-level materials science problems, including those that are particularly challenging and require advanced reasoning and computation.

For reference, we provide the original problem numbers as they appear in the textbooks. For each problem, we provide the correct answer and incorrect choices, with a way of making incorrect choices. The information is available on HuggingFace [16].

## 2.3 Creating Problem Sentences

After selecting problems for the benchmark, each was manually converted into a text document. Many problems required modifications before they could be used as input, such as those that referenced values in a separate table or that lacked specified units for answers. To address this, we made necessary adjustments to the problems by adding the appropriate tables or specifying units, without altering the systems they address or their original intent. Since the unit of answer is specified in the problem sentence, the correct answer does not include the unit. The answers include not only numerical values but also textual responses, such as electron configurations, crystal orientations, magnitude relations of numerical values, or decrease/increase in properties. For multiple-choice problems, the problem sentences were created by appending a choice list to the corresponding free-response problems. For example, "Choose the answer from the following multiple-choices: (a) Sn: 27.5at%, Pb: 72.5at%, (b) Sn: 80.9at%, Pb: 19.1at%, (c) Sn: 46.3at%, Pb: 53.7at%, and (d) Sn: 72.5at%, Pb: 27.5at%." was added after "What is the composition (Sn at%, Pb at%) of an alloy that contains 98 g tin and 65 g lead?".

This set of problem-answer pairs, created through these methods, serves as the benchmark dataset.

## 3 Experiments

### 3.1 Experiment Setup

We evaluated ChatGPT-3.5 and ChatGPT-4 [17], Bard (at the time of experiments, now Gemini) [18], and GPT-3.5 and GPT-4 via the OpenAI API [19] using our benchmark dataset.

Because CoT is automatically included in three models, we did not use any specific prompts when inputting problems into ChatGPT-3.5, ChatGPT-4, and Bard. When problems were input using the OpenAI API, a prompt was sometimes used, which will be mentioned in the results section. We did not manually enable any plug-ins, though Python is automatically integrated with ChatGPT-4.

For the free-response problems, responses were obtained by directly inputting the problem statements into the conversational interfaces of ChatGPT and Bard between 16 Nov. and 7 Dec., 2023. Regarding the multiple-choice problems, two primary methods were used to obtain responses. The first involved directly inputting problem statements modified for multiple-choice problems mentioned above into the conversational interfaces of ChatGPT and Bard between 18 Dec., 2023 and 7 Jan., 2024. The second method involved using the OpenAI API, which allowed responses for the same problem to be repeatedly obtained as new chats automatically generated by Python. On the conversational interface of ChatGPT, users could select only either ChatGPT-3.5 or ChatGPT-4 (at the time, now also ChatGPT-4o) as the model. However, the OpenAI API allowed users to choose different versions of ChatGPT-3.5, such as gpt-3.5-turbo-0125 and gpt-3.5-turbo-1106, or ChatGPT-4, including gpt-4-0125-preview, gpt-4-1106-preview and gpt-4 [20].

For both free-response and multiple-choice problems, the correctness of the responses was manually conducted by an expert.

3.2 Results and Analysis

3.2.1 Overall Results

The overall accuracy of all experiments is summarized in Table 1, along with the dates when the problem statements were input, the model versions used, and whether the system messages were employed.

The free-response questions were input twice separately into ChatGPT-3.5 and Bard. The accuracies listed in Table 1 for these response, 0.28 and 0.30 for ChatGPT-3.5 and Bard respectively, are the average values of the two inputs. Due to the accuracy deviation between the first and second inputs, there appears to be no significant difference between 0.28 for ChatGPT-3.5 and 0.30 for Bard. However, the disparity between 0.28 for ChatGPT-3.5 and 0.64 for ChatGPT-4 is more than double, indicating a significant difference.

Responses for multiple-choice problems with ChatGPT-3.5 and ChatGPT-4 were obtained once per problem, while responses with Bard were obtained six times per problem. The scores in Table 1 are the averaged values. For Bard, three responses were obtained at one time for each problem.

Occasionally, despite the problems being multiple-choice, the model responded with "there are no appropriate answers in the given choices" instead of selecting a choice. In such cases, these responses were considered incorrect.

Responses to the multiple-choice problems were also obtained using OpenAI's API, with ten responses per problem. Different versions of GPT-3.5 and GPT-4 were used. Since some responses did not select any of the given choices with ChatGPT-3.5 and ChatGPT-4, a system message "You are a helpful assistant and answer user's question with multiple-choices by giving one correct choice" was added before each problem statement for several GPT-3.5 and GPT-4 models, as indicated in the "system message" row in Table 1.

For the results obtained using OpenAI's API, the accuracy scores in Table 1 are divided by '/' for two different accuracy definitions. The first score, before '/', represents the accuracy determined by taking the most frequent response out of 10 as the answer (majority vote) from the GPT models. The second score, after '/', is the accuracy calculated by averaging the correct answer rate for each problem, which is based on the number of correct answers out of 10. If the responses for a problem are "a, a, b, c, c, b, c, b, b, a" and the correct answer is "b", the majority vote would be "b", making it the correct response, while the correct answer rate would be 0.4. The latter score is always lower than the former. The latter score should be regarded as accuracy statistically. However, the former could be considered to simulate the accuracy of a model's response (probability of outputting a correct answer). This discrepancy is illustrated in the histograms in Fig. 1. The problems were divided into eleven classes based on the correct answer rate (abscissa in the figure), and the number of problems with correct (orange) / incorrect (blue) answers for each class was counted, which is the ordinate of Fig. 1 (b) and (d) for GPT-3.5 and GPT-4, respectively. The number of problems with correct / incorrect answers was converted to the percentage of correctness (or incorrectness), which is shown in Fig. 1 (a) and (c) for GPT-3.5 and GPT-4, respectively. For GPT-3.5, it is evident that even when the correct answer rate is 0.4, the probability of a correct answer is more than 0.8. Therefore, the "majority vote" method appears to yield higher accuracy.

3.2.2 Comparison among Different Models on Free-Response Problems

Figure 2 provides a comparison of the correct/incorrect response patterns across different models for free-response problems. Each small square represents a correct/incorrect response for problem #1 - #144, arranged sequentially from left to right, with the right edge continuing to the left of the second row, and so forth. Pink color shows a correct answer, while black represents an incorrect answer. As seen from the figure, the response patterns for the different models are similar, though ChatGPT-4 shows a higher number of correct answers. There are 45 problems out of 144 (31%) where all three models provided either correct or incorrect answers. The results for 81 problems out of 144 (56%) are consistent between ChatGPT-3.5 and Bard.

3.2.3 Comparison among Different Models on Multiple-choice Problems
Figure 3 illustrates the correct/incorrect response patterns for different models on multiple-choice problems, using the same style as Fig. 2. Similar to free-response problems, the patterns among the three models are comparable. For 43 problems out of 164 (26%), all three models either provided correct or incorrect answers. This level of agreement is notably high compared to free-response problems (31%), particularly given that Bard's responses were analyzed six times. This high degree of consistency might be attributed to the fact that 101 problems out of 164 (62%) were either correct or incorrect across all six responses from Bard.
The results for 29 problems out of 164 (18%) were consistent between ChatGPT-3.5 and Bard. The agreement between ChatGPT-3.5 and Bard on multiple-choice problems was significantly lower compared to free-response problems. This difference is likely due to Bard's six responses being analyzed for multiple-choice problems, compared to only two for free-response problems. Additionally, the results for 14 problems out of 164 (9%) were consistent between ChatGPT-4 and Bard. This degree of agreement is significantly lower than that between ChatGPT-3.5 and Bard, suggesting that the performance of Bard is closer to that of ChatGPT-3.5 rather than ChatGPT-4

3.2.4 Comparison between Free-Response vs. Multiple-choice Problems
Figure 4 shows that the patterns of correct/incorrect answers for all models (ChatGPT-3.5, ChatGPT-4 and Bard) are quite similar for free-response and multiple-choice problems. Problems that were not in free-response format are represented by white squares in the free-response column of the figure. All models failed to provide a correct answer for 12 of these problems. When comparing the performance on free-response and multiple-choice problems for each model, the overall pattens are similar. However, there are occasional discrepancies: some problems were answered correctly in the free-response format but incorrectly in the multiple-choice format, and vice versa. These differences result in only slight variations in the percentage of correct answers between the free-response and 4-choice formats.

3.2.5. Comparison between Direct Input and OpenAI API (Multiple-choice Problems)

Comparing results between direct input and OpenAI API under the same conditions (gpt-3.5-turbo-1106 and gpt-4-1106-preview, both without system messages), the OpenAI API showed a slightly higher accuracy for both GPT3.5 (0.33 vs. 0.39) and GPT4 (0.66 vs. 0.70). The experiments using OpenAI API were conducted on April 18, after the release of gpt-4-turbo-2024-04-09 model. The models used in these experiments were the default models available at the time of the direct input experiments, though it is unknown if there were any minor updates on older models after the release of the newer ones.

Figure 5 compares the correct/incorrect response patterns between direct input and OpenAI API for GPT-3.5 (Fig. 5(a)) and for GPT-4 (Fig. 5(b)). For the OpenAI API, the correct response is derived by a majority vote, where the most frequent answer out of 10 is considered correct for each problem. White squares in the figures represent cases where multiple choices received the highest number of votes. For example, if the correct answer is (a) but the responses include 5-times (a) and 5-times (c), a white square is used. The percentage of correct answers using majority vote is much higher for both GPT3.5 (0.33 vs. 0.43) and GPT4 (0.66 vs. 0.81). This means there were problems that were incorrect with direct input but correct with the OpenAI API, although the general patterns are quite similar for both GPT-3.5 and GPT-4. When comparing GPT-3.5 and GPT-4, the patterns are similar, but GPT-4 shows a significantly higher percentage of correct answers. It should be noted that there are no problems that were correct with GPT-3.5 for both direct input and OpenAI API but incorrect with GPT-4 for both formats. This suggests that problems that can be solved with GPT-3.5 are expected to be solved correctly by GPT-4 when majority voting is applied. There are no significant differences in the characteristics between GPT-3.5 and GPT-4 beyond performance. Both models had 18 problems that were incorrect for both direct input and OpenAI API.

3.2.6 Effect of the System Message

A significant number of responses failed to provide an answer with a choice, despite the problem instructions specifying the need for multiple-choice responses.

To address this, we incorporated the system statement "You are a helpful assistant and answer user's question with four choices by giving one correct choice." using the OpenAI API. However, this adjustment resulted in a lower percentage of correct answers compared to direct input for both GPT-3.5 (0.43/0.39 for direct input vs. 0.36/0.32) and GPT-4 (0.81/0.70 for direct input vs. 0.54/0.53), even with the updated models (gpt-3.5/4-*-0125-*), as shown in Table 1. When comparing responses with system messages between gpt-4-0125-preview and gpt-4-0613, the older model, gpt-4-0613 (this is a model in 2023, while gpt-4-0125-preview is in 2024), demonstrated lower scores.

Figures 6 (a) and 6 (b) illustrate the patterns of correct/incorrect responses for GPT-3.5 and GPT-4, respectively, both with and without system messages. While there is some similarity, beyond randomness, between the response patterns for both models, no overlap was observed in problems that were incorrect without the system message but correct with it. This implies that the system message had little impact, primarily leading to a decrease in accuracy. Figure 7 shows histograms comparing incorrect (blue) and correct (orange) answers percentage (majority vote) against correct answer rate, similar to Fig. 1, but focusing on the effect of the system message for GPT-3.5. This figure indicates that responses with the system message were more random, with a lower percentage of correct answers, especially with a correct answer rate of 0.4.

The impact of the system message varied depending on the problem. For instance, with gpt-3.5-turbo-0125, one problem was "Calculate a planar density value ($m^{-2}$) for (110) plane for vanadium." (3-55). Without the system message, all 10 responses used CoT and averaged 191 words. In contrast, with the system message, CoT was not used, and responses were simply "The correct choice is (*) .", where * is either a, b, c, or d.

Another example involved "Calculate the expected 2-theta diffraction angle (degrees) for the first-order reflection from the (113) set of planes for FCC platinum when monochromatic radiation of wavelength 0.1542 nm is used." (3-59). Occasionally, CoT was used with the system message. Without the system message, all 10 responses used CoT and averaged 220 words. With the system message, CoT was used in 4 out of 10 responses, which averaged 173 words. The probability of using CoT when the system message was used varied with the problem. When the number of words in CoT without the system message was small, the probability of using CoT with the system message was low. Some problems saw minimal use of CoT even without system messages.

It should be noted that this analysis of the system message's effect is specific to this particular message with these problems and may not generalize.

4 Error Analysis

Due to the limited size of the benchmark dataset and the intentional avoidance of repetitive problems, a statistical analysis of errors is not feasible. Instead, this section focuses on identifying differences and similarities among the models based on observed error patterns.

4.1 Difference between ChatGPT-3.5 or Bard and ChatGPT-4

As seen in Fig. 2, 3, and 4, the pattens of correct/incorrect responses are generally similar among the three models with direct input. However, ChatGPT-4 consistently outperforms both ChatGPT-3.5 and Bard in terms of accuracy. The following analysis explores the reasons behind the lower accuracy of ChatGPT-3.5 and Bard compared to ChatGPT-4.

For free-response problems, there were 35 problems where ChatGPT-4 provided correct answers while ChatGPT-3.5 and Bard did not. The predominant errors made by ChatGPT3.5 and Bard were due to miscalculations, especially involving factorial or exponent functions. ChatGPT-4's integration with Python plugins helps mitigate such errors. Additionally, ChatGPT-3.5 and Bard struggled with deriving equations using parameters, such as 'r=($\sqrt{2}$-1)R'. Another common issue involved the use of inappropriate equations for calculations, and some errors were attributed to difficulties in understanding the problem statements.

For multiple-choice problems, there were 29 problems where ChatGPT-4 correctly answered questions that ChatGPT-3.5 and Bard did not. Similar to free-response problems, miscalculations—especially those involving factorials or exponents—were a major source of errors for ChatGPT3.5 and Bard. Additionally, both models struggled with understanding problem statements and deriving the correct equations.

4.2 Problems that All Models Failed to Give Correct Answers

There were 11 problems where all models failed to provide a correct answer, regardless of whether the problems were free-response or multiple-choice. There were 18 problems where both GPT-3.5 and GPT-4, using direct input and OpenAI API, produced incorrect answers for multiple-choice problems. The overlap of 11 and 18 problems, which means that regardless of direct input or OpenAI API, free-response or multiple-choice problem, no models could give correct answers, consisted of 9 problems. Among them, relatively straightforward errors (3 problems) were due to models misunderstanding the problem statements. One example is "Calculate the number of atoms for an HCP unit cell. Choose the answer from the following four choices: (a) 2, (b) 6, (c) 3, and (d) 4", the correct answer is (a) 2. However, the models failed to consider "unit cell" and instead seemed to calculate for a hexagonal prism. Another 4 problems appeared challenging due to the models' difficulty in following the necessary logical steps. For example, consider the problem "Boron atoms are to be diffused into a silicon wafer using both predeposition and drive-in heat treatments; the background concentration of B in this silicon material is known to be $1 \times 10^{20}$ atoms/m3. The predeposition treatment is to be conducted at 900°C for 30 minutes; the surface concentration of B is to be maintained at a constant level of $3 \times 10^{26}$ atoms/m3. Drive-in diffusion will be carried out at 1100 °C for a period of 2 h. For the diffusion coefficient of B in Si, values of $Q_d$ and $D_0$ are 3.87 eV/atom and $2.4 \times 10^{-3}$ m2/s, respectively. Calculate the value of $x_j$ (μm) for the drive-in diffusion treatment. Choose the answer from the following four choices: (a) 1.08, (b) 1.22, (c) 1.27, and (d) 2.16.". In this case, the models applied the diffusion equation for the 1100 °C drive-in treatment without accounting for the concentration changes during the predeposition phase.

5 Conclusion

In conclusion, this paper introduces MaterialBENCH, a college-level dataset focused on materials science problems. The dataset features two types of problems: free-response and multiple-choice, with the primary distinction being the format of the answers. We conducted experiments with MaterialBENCH using ChatGPT-3.5, ChatGPT-4 and Bard models, and both GPT-3.5 and GPT-4 via the OpenAI API. The findings of this study highlight several key points: (1) GPT-3.5-based models and Bard exhibit notable weaknesses in solving problems that require mathematical calculations, (2) there are distinct similarities and differences in problem-solving capabilities between GPT-4-based models and those based on GPT-3.5-based or Bard, (3) solving complicated logical problems remains challenging for all models. We envision that the MaterialBENCH benchmark dataset will serve as a valuable resource for future research, contributing to a deeper understanding and improvement of domain-specific LLMs' problem-solving capabilities in materials science.

Table 1 Summary of overall accuracy for all experiments with MaterialBENCH.

|  |  | Chat3.5 | GPT3.5(API) | Chat4 | GPT4(API) | Bard |
|---|---|---|---|---|---|---|
| free | dates | 2023.11.16 |  | 2023.11.16 |  | 2023.11.17 |
|  | model | gpt-3.5-turbo-1106 |  | gpt-4-1106-preview |  |  |
|  | system statements | none |  | none |  | none |
|  | score | 0.28 |  | 0.64 |  | 0.30 |
| choice | dates | 2024.1.4 | 2024.4.18 | 2023.12.30-2024.1.7 | 2024.4.18 | 2023.12.28-12.30 |
|  | model | gpt-3.5-turbo-1106 | gpt-3.5-turbo-1106 | gpt-4-1106-preview | gpt-4-1106-preview |  |
|  | system statements | none | none | none | none | none |
|  | score | 0.33 | 0.43/0.39 | 0.66 | 0.81/0.70 | 0.34 |
|  | dates |  | 2024.3.4-4.2(gpt-3.5-turbo) |  | 2024.4.11(gpt-4-turbo-preview) |  |
|  | model |  | =gpt-3.5-turbo-0125 |  | =gpt-4-0125-preview |  |
|  | system statements |  | used |  | used |  |
|  | score |  | 0.36/0.32 |  | 0.54/0.53 |  |
|  | dates |  |  | 2024.4.15-16 | 2023.3.4-4.2(gpt-4) |  |
|  | model |  |  | gpt-4-turbo-2024-04-09 | =gpt-4-0613 |  |
|  | system statements |  |  | none | used |  |
|  | score |  |  | 0.79 | 0.41/0.40 |  |

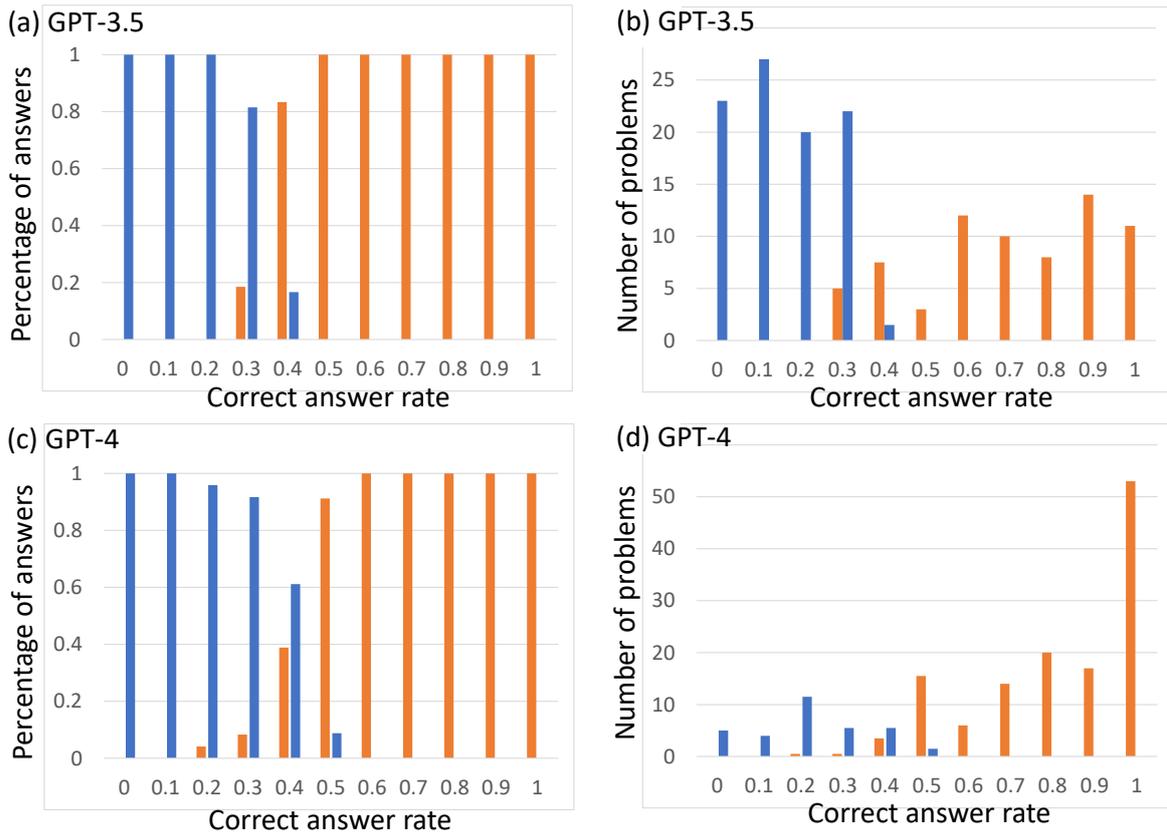

Fig. 1 Histograms of correct answer rate vs. number of incorrect (blue) and correct (orange) answers by majority vote (b) and (d), vs. percentage of incorrect and correct ones (a) and (c).

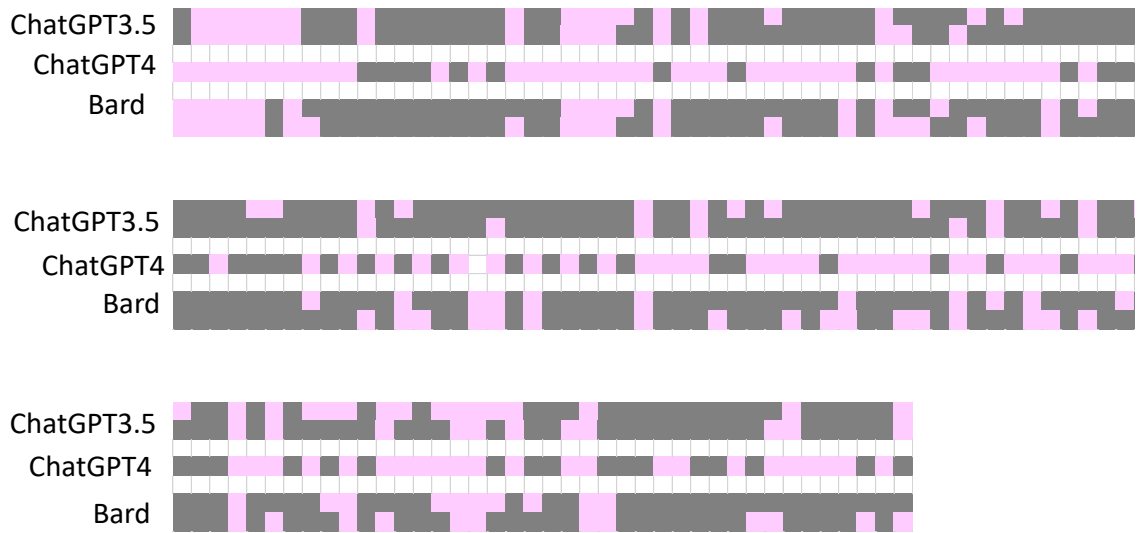

Fig. 2 Correct/incorrect patterns of answer for free-response problems with different models.

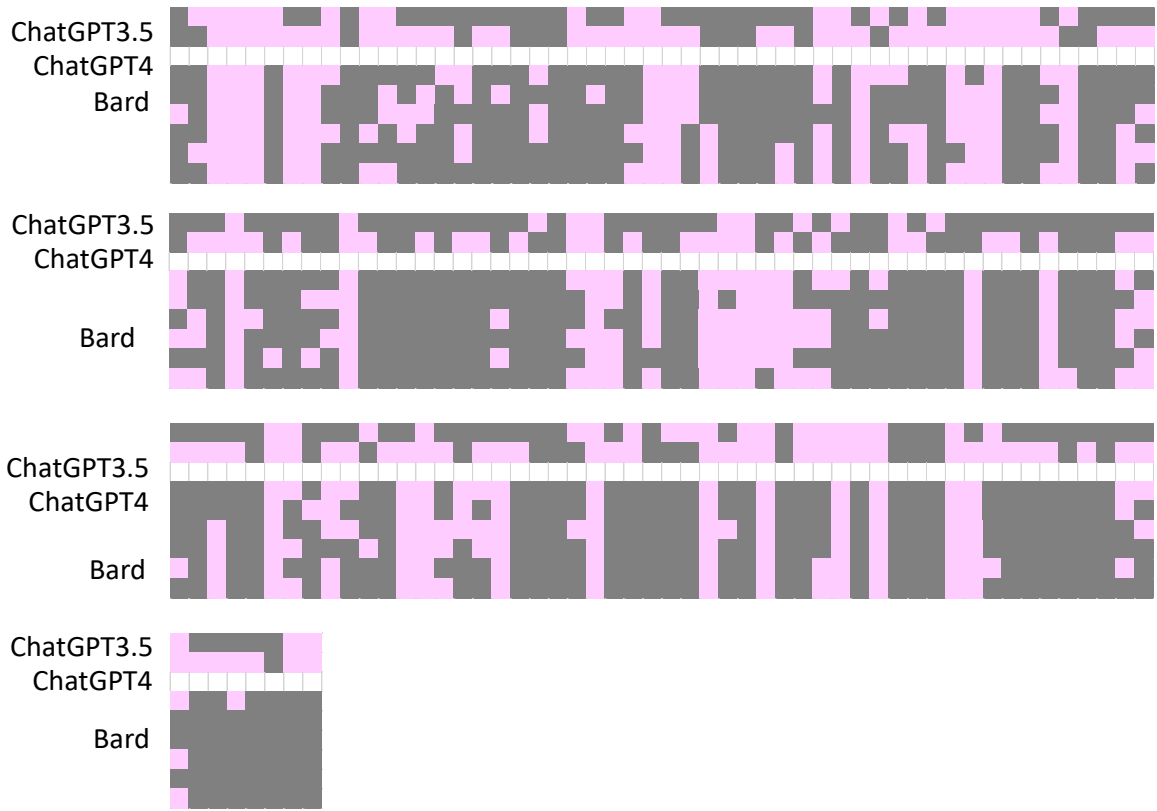

Fig. 3 Patterns of correct/incorrect response on multiple-choice problems with different models.

Fig. 4 Comparison of correct/incorrect patterns between free-response and multiple-choice problems for all models.

(a)

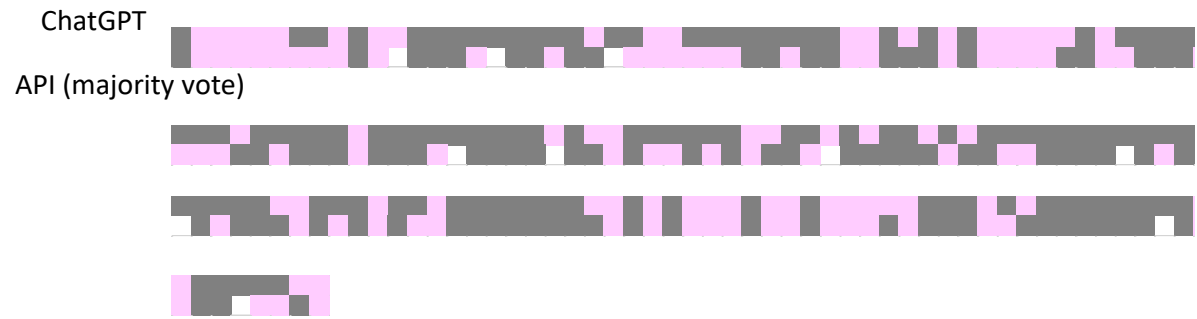

(b)

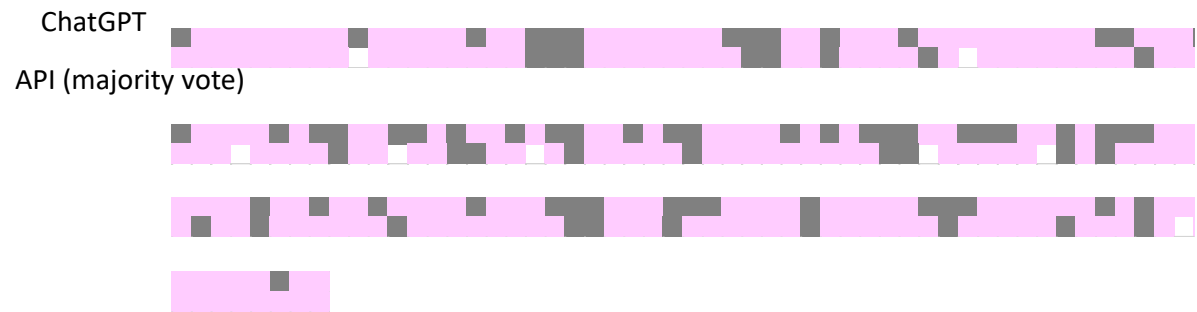

Fig. 5 Comparison between correct/incorrect pattens of responses with direct input (ChatGPT) and OpenAI-API for GPT-3.5-base (a) and GPT-4-base (b).

(a) GPT-3.5
without system message (1106)
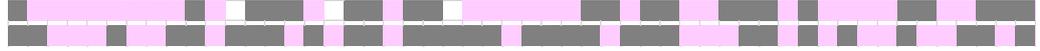
with system message (0125)
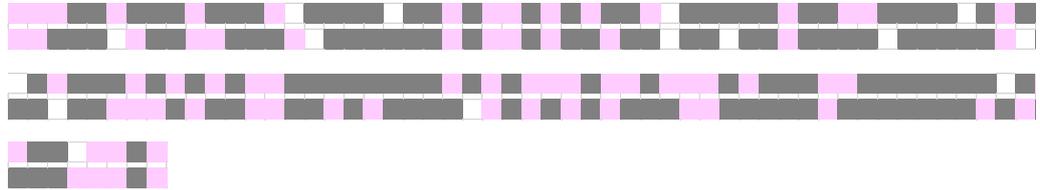

(b) GPT-4
without system message
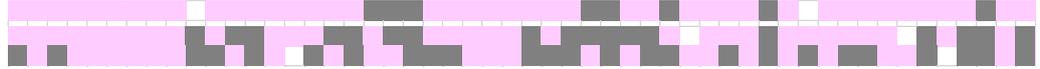
with system message (0125: upper, 0613: lower)
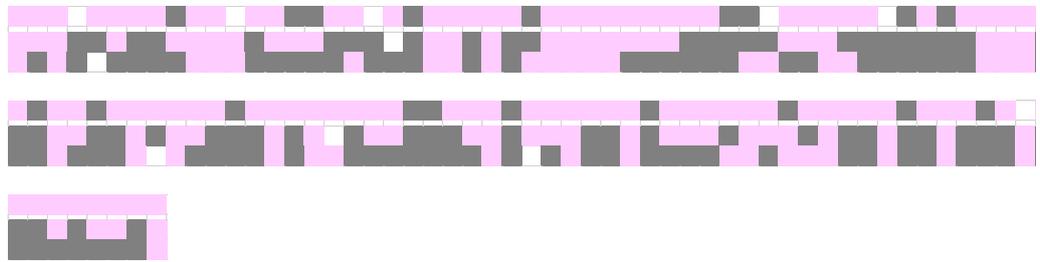

Fig. 6 Comparison between correct/incorrect pattens of responses without and with system message for GPT-3.5 (a) and GPT-4 (b).

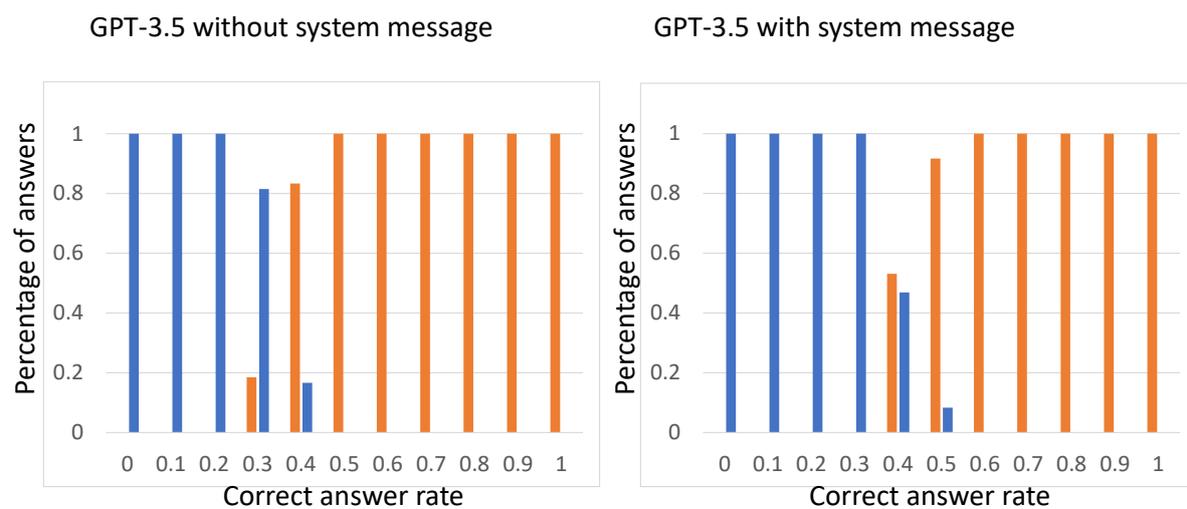

Fig. 7 Comparison of histograms of correct answer rate vs. incorrect (blue) and correct (orange) answers by majority vote, between responses without and with system message for GPT-3.5.